\documentclass[letterpaper, 10 pt, conference, twoside]{IEEEtran}

\IEEEoverridecommandlockouts                              

\usepackage{xcolor}
\usepackage[pdftex]{graphicx}
\usepackage{subfigure}
\usepackage{amssymb}
\usepackage{amsmath}
\usepackage{bm}
\usepackage{hyperref}
\usepackage{makecell}
\usepackage{xcolor}
\usepackage[linesnumbered,ruled,vlined]{algorithm2e}
\usepackage{algpseudocode}
\usepackage{subfiles}
\usepackage{siunitx}

\DeclareMathOperator{\trace}{tr}
\DeclareMathOperator{\sign}{sgn}

\newcommand{\m}[1]{\ensuremath{#1}}
\newcommand{\T}{\intercal}
\newcommand{\reals}{\mathbb{R}}
\newcommand{\s}[1]{\ensuremath{\mathcal{#1}}}

\newcommand{\iangle}{\gamma}
\newcommand{\mparams}{\m{w}}
\newcommand{\etal}{\emph{et~al.}}

\SetCommentSty{mycommfont}
\SetKwInput{KwInput}{Input}                
\SetKwInput{KwOutput}{Output}              

\begin{document}

\title{\LARGE \bf
Self-Supervised Depth Correction of Lidar Measurements from Map Consistency Loss}

\author{Ruslan Agishev, Tomáš Petříček, Karel Zimmermann
\thanks{Manuscript received: February, 27, 2023; Revised May, 13, 2023; Accepted June, 12, 2023.}
\thanks{This paper was recommended for publication by
Editor Javier Civera upon evaluation of the Associate Editor and Reviewers’ comments.}
\thanks{This work was co-funded by the Europen Union under the project Robotics and advanced industrial production (reg. no. CZ.02.01.01/00/22\_008/0004590), by Czech Science Foundation under Project 20-29531S, by OP VVV MEYS funded project CZ.02.1.01/0.0/0.0/16\_019/0000765 ``Research Center for Informatics'', and by the European Union's Horizon Europe Framework Programme under the euRobin project (ID: 101070596).
R.~Agishev and T.~Petříček were supported by 
Grant Agency of the CTU Prague under Project SGS22/111/OHK3/2T/13.
\emph{(Corresponding author: R.~Agishev.)}}
\thanks{The authors are with the Department of Cybernetics, Faculty of Electrical Engineering, Czech Technical University in Prague, 166 36 Prague, Czech Republic (e-mail: agishrus@fel.cvut.cz; tpetricek@gmail.com; zimmerk@fel.cvut.cz)}
\thanks{Digital Object
Identifier (DOI): see top of this page.}}

\markboth{IEEE Robotics and Automation Letters. Preprint Version. Accepted June, 2023}
{Agishev \MakeLowercase{\textit{et al.}}: Self-Supervised Depth Correction of Lidar Measurements}

\maketitle

\begin{abstract}

Depth perception is considered an invaluable source of information in the context of 3D mapping and various robotics applications.
However, point cloud maps acquired using consumer-level \emph{light detection and ranging} sensors (lidars) still suffer from bias related to local surface properties such as measuring beam-to-surface incidence angle.
This fact has recently motivated researchers to exploit traditional filters, as well as the deep learning paradigm, in order to suppress the aforementioned depth sensors error while preserving geometric and map consistency details.
Despite the effort, depth correction of lidar measurements is still an open challenge mainly due to the lack of clean 3D data that could be used as ground truth.
In this paper, we introduce two novel point cloud map consistency losses, which facilitate self-supervised learning on real data of lidar depth correction models.
Specifically, the models exploit multiple point cloud measurements of the same scene from different view-points in order to learn to reduce the bias based on the constructed map consistency signal.
Complementary to the removal of the bias from the measurements, we demonstrate that the depth correction models help to reduce localization drift.
Additionally, we release a data set that contains point cloud scans captured in an indoor corridor environment with precise localization and ground truth mapping information~\footnote{\url{http://ptak.felk.cvut.cz/vras/data/fee_corridor/}}.

\end{abstract}


\section{Introduction}



\begin{figure}[t]
  \centering
  \includegraphics[width=\columnwidth]{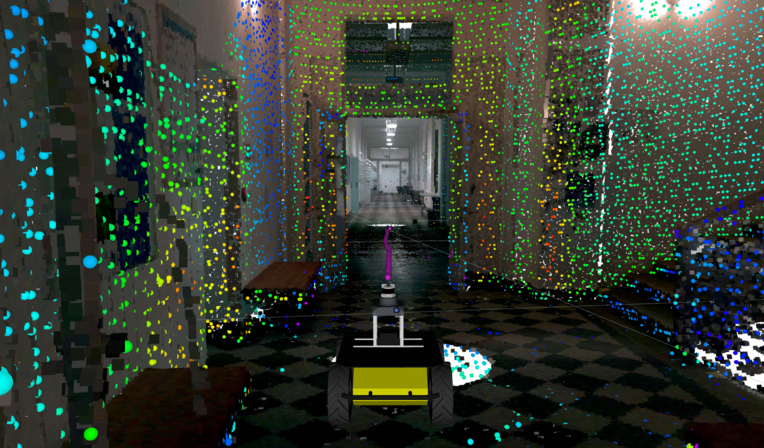}
  \caption{Mobile robot with lidar mapping a corridor-like environment. Points color denote lidar beams incidence angle (darker color - higher angle). Measurements with higher incidence angle have higher bias \cite{Laconte-2019-ICRA}. The snapshot is taken from the FEE Corridor data set (Sec.~\ref{subsection:dataset}). The data set is recorded for learning the lidar measurements correction models as well as to evaluate SLAM and reconstruction algorithms.}
  \label{fig:demo_husky_fee_corridor_inc_angles}  
\end{figure}

Lidars has been deployed in a myriad of applications, including but not limited to \emph{simultaneous localization and mapping} (SLAM) for mobile robots and autonomous cars, and remote sensing ~(\cite{NiMeister-2001-TGRS, Wulder-2012, Sun-2000-TGRS}). 
However, point cloud maps acquired using consumer-level \emph{light detection and ranging} sensors (lidars) still suffer from various biases induced by immediate capturing conditions. It is well-known that different environmental conditions such as illumination, temperature or humidity~\cite{heinzler2019weather} and surface reflectance~\cite{li2015remote} affect lidar performance. A lidar measurements accuracy could also be influenced by diverse material types and scanned distance (see and example in~\autoref{fig:point-to-plane_inc_angle}). In this work we confirm that a substantial bias typically emerges when measuring surfaces with a high incidence angle (\autoref{fig:demo_husky_fee_corridor_inc_angles}). The resulting measurement bias then negatively influences both the localization and mapping accuracy; therefore, some kind of bias-compensation module is highly desirable.

\begin{figure}
    \centering
    \includegraphics[width=\columnwidth]{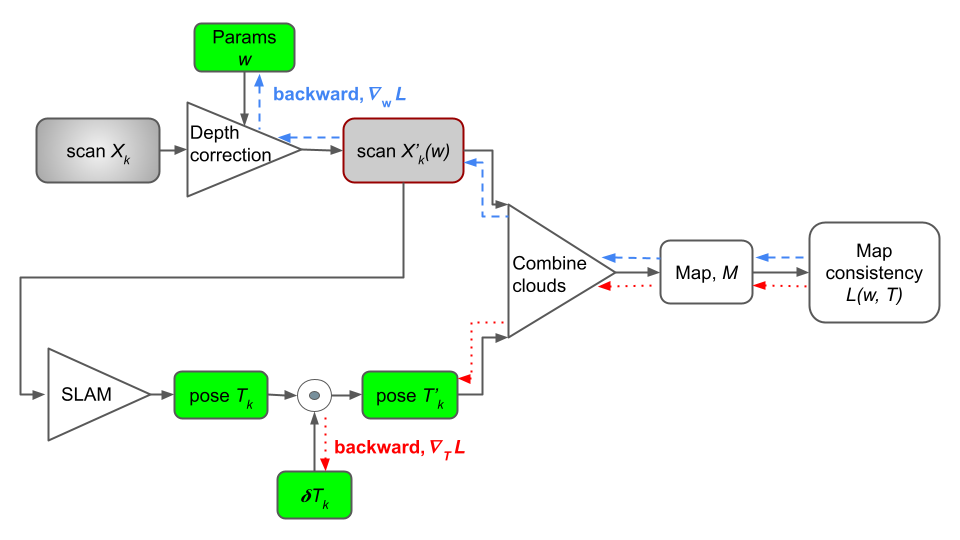}
    \caption{Self-supervised pipeline to learn lidar measurements correction model and optimize sensor poses. Point cloud map is constructed from corrected lidar measurements and sensor poses. We compute the map consistency loss on the resultant map and propagate the error back to model parameters as well as to corresponding poses correction terms. The depth correction model removes the bias from raw lidar scans.}
    \label{fig:depth_corr_scheme}
\end{figure}

In many applications, such as the outdoor deployment of a mobile robotic platform, the accurate capturing conditions are apriori unknown and dynamically evolving. In such situations, the bias-compensating module has to be regularly (or even continuously) adapted on immediate capturing conditions. Since reliable measuring of exact capturing conditions would require various exotic sensors onboard, we instead focus on a self-contained solution independent of other sensor modalities that can naturally adapt the bias-compensating module just from lidar measurements. Since there are no ground truth depths that can be used for supervised adaptation, an alternative source of self-supervision is required.
%

In the task of lidar-based SLAM, the natural expectations
are that the resulting map will look like a map of a typical environment. 
In order to perform the adaptation, we address the two underlying sub-problems: (i) What is a suitable self-supervised loss that measures the inconsistency between reality and expectations?  (ii) What is the structure of the depth-correcting module that allows for end-to-end learning from this self-supervised loss?

Towards this end, we propose the fully-differentiable depth-correcting module, which leverages a state-of-the-art first-principle model of lidar-surface interaction. We also introduce a so-called \emph{map consistency loss} - the measure that encodes our prior assumptions about the appearance of the typical environment. There is a trade-off between the strength of the prior assumptions and the universality of the solution: the more substantial the assumptions, the better the self-supervision, but the lower the variability of environments the robot can operate in. Since we target on a generally applicable solution, we introduce weaker assumptions that are satisfied in an almost arbitrary environment.
We experimentally demonstrate that even these weak assumptions are sufficient for the successful self-supervised learning of the bias/depth-compensating module since the resulting approach increases the localization and mapping accuracy of a typical SLAM method.


The \textbf{key contributions} of the work are following.
\begin{itemize}
    \item Novel point cloud map consistency losses are designed, which could be used in a self-supervised manner for depth measurements correction, as well as sensors calibration tasks.
    
    \item A self-supervised training method is introduced aimed to learn lidar measurements correction models.
    
    \item It is demonstrated that utilization of the depth correction models not only allows to suppress the bias from lidar measurements, but also to create more accurate maps, and reduce localization drift in SLAM scenarios.

    \item Novel data sequences are released which contain accurately localized point cloud data with a ground truth map.
\end{itemize}

The paper is organized as follows.
Section~\ref{sec:related_work} presents related work.
The description of depth correction models and map consistency loss is given in Section~\ref{sec:methodology}.
Experiments and evaluation studies are described in Section~\ref{sec:experiments}.
Finally, conclusions are drawn in Section~\ref{sec:conclusion}.
To facilitate reproducibility of the paper, we provide the source code%
\footnote{\url{https://github.com/ctu-vras/depth\_correction}}.

\section{Related Work}~\label{sec:related_work}
A laser beam width is typically modeled as a Gaussian function of the distance along the beam~\cite{Bandres:04}.
Using non-zero-mean Gaussian beam model, Laconte \etal~\cite{Laconte-2019-ICRA} were able to measure the bias of several commercial lidar sensors as a function of depth and incidence angle.
The authors additionally present an experimental methodology on estimation of the bias model parameters for a lidar sensor.
We provide an extension of their work, by introducing a learning method to estimate the bias model parameters using raw sensory measurements.
Sterzentsenko \etal~\cite{sterzentsenko2019self} presented self-supervised depth denoising method for RGB-D sensors.
Similar to our approach, the authors support the idea of utilizing spatial consistency and surface smoothness as optimization objectives.
In our work, the lidar depth correction model is used that was introduced in \cite{Kuemmerle-2020-ICRA}.
It estimates the bias as a polynomial function of laser beam incidence angle, Section~\ref{sec:depthcorr_model}.
Additionally, we propose a new model that includes dependence of the bias on measured distance by a laser beam (Section \ref{sec:depthcorr_model}). This idea is motivated by the depth measurements correction results presented in \cite{Laconte-2019-ICRA}.
The authors of \cite{Kuemmerle-2020-ICRA} utilize the model for lidars and cameras calibration scenario. However, the application of the model in localization and mapping scenarios was not studied in their paper.

As we pose the task of global point cloud map consistency estimation as optimization problem, a differentiable loss function is required as a minimization objective.
The Chamfer loss introduced in \cite{fan2017point} is one of the most famous differentiable function for point sets.
It is usually utilized to evaluate a distance between two point clouds, thus is typically suitable for supervised tasks.
It was shown in \cite{agarwal2019learning} that the introduced quadric loss in combination with the Chamfer loss improves 3D reconstruction accuracy by preserving sharp edges of scene objects. The loss is designed to reduce quadric error between the reconstructed points and the input surface.
However, as it was mentioned by the authors of \cite{agarwal2019learning}, the quadric loss can not preserve the input point distribution, since the quadric loss is zero anywhere on the plane.
On the contrary, in order to reduce the lidar measurements bias caused by high incidence angle, we target to correct mainly the points which belong to planar surfaces.
Therefore a designed global map consistency loss should be able to encourage points from different scans that represent such objects as walls, floors, and roads to belong to the same plane.
Moreover, as shown in the next Section~\ref{sec:methodology}, we do not require ground-truth data to evaluate the point cloud consistency, and the introduced losses are computed directly on the global map independently on the number of lidar scans and view-points forming the input point cloud.

\section{Methodology}\label{sec:methodology}
The method is outlined in \autoref{fig:depth_corr_scheme}.
Briefly, a depth correction model is applied to individual laser scans (point clouds).
The corrected point clouds are transformed into a global reference frame (that of the first measurement) to build a global point cloud map.
Then, a consistency loss is computed for this map and its gradient is propagated back in the computation graph to update model learnable parameters as well as to correct sensor poses.
We discuss further data representation, models choice, optimization algorithm as well as map consistency functions used as optimization criteria.

\subsection{Depth Correction Model}~\label{sec:depthcorr_model}

Lidar systems are based on laser ranging, which measures the distance between a sensor and a target based on half the elapsed time between the emission of a pulse and the detection of a reflected return~\cite{Baltsavias-1999}.
The systems are classified as either discrete return (record single or multiple returns from a given laser pulse) or full waveform recording (digitize the entire reflected energy)~\cite{Wulder-2012}.
In discrete return model, as the laser signal is reflected back to the sensor, large peaks, (i.e., bright returns), are interpreted as objects in the path of the beam.
This makes them favourable for 3D mapping applications, where the sensed environment is represented by clouds of points characterising intercepted features.
For the real lasers, the light beam width is not constant along its propagation axis~\cite{Svelto-2010-Lasers}.
That is why the intensity distribution is not uniform along a beam direction.

When  the  laser  beam  hits  an  oriented  surface, a  great  amount  of intensity returns sooner to the sensor than if the surface was not oriented.
This means that the returned waveform will have a different shape depending on the incidence angle, reflectance, and distance  inducing a bias in the measurement~\cite{Laconte-2019-ICRA}. Since we focus on a self-contained system, which will be independent of other sensors that would allow to predict surface properties directly,
the proposed depth-correcting module infer the immediate bias based on the incidence angle and measured depth.

\begin{figure}[t]
  \centering
  \includegraphics[width=\columnwidth]{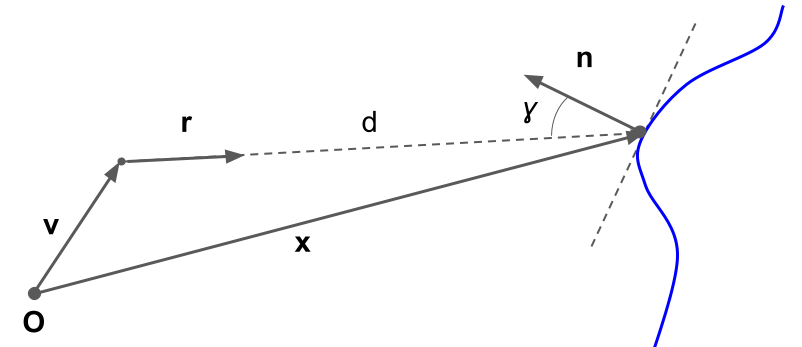}
  \caption{Lidar scan points representation. A ray represented by vector $\mathbf{r}$ falls at an oriented surface (drawn as a blue curve); $\mathbf{v}$ is the vector pointing to the ray origin (sensor location), $\mathbf{x}$ is a vector defining the measured point location. The surface normal $\mathbf{n}$ at the beam heating point forms the lidar ray incidence angle $\gamma$ with the vector $-\mathbf{r}$.}
  \label{fig:depth_cloud_notations}
\end{figure}

Lidar measurement (a scan) is represented as a set of points $\mathbf{x}_i \in \s{X} \subset \reals^{3 \times 1}$.
Each point $\mathbf{x}$ is expressed as
\begin{equation}
    \mathbf{x} = \mathbf{v} + d \mathbf{r},
\end{equation}
in terms of origin $\mathbf{v} \in \reals^{3 \times 1}$ of the measuring ray, its direction $\mathbf{r} \in \reals^{3 \times 1}$, $\|\mathbf{r}\| = 1$, and depth $d > 0$,~\autoref{fig:depth_cloud_notations}.
For normal vector $\mathbf{n} \in \reals^{3 \times 1}$, $\|\mathbf{n}\| = 1$, which is perpendicular to the surface at point $\mathbf{x}$ and pointing outward from the object, incidence angle is given by
\begin{equation}
\iangle = \arccos(-\mathbf{n}^\T \mathbf{r}).
\end{equation}

Following~\cite{Laconte-2019-ICRA}, we estimate depth bias $\epsilon$ from measured depth $d$ and incidence angle $\iangle$, resulting in corrected depth
\begin{equation}
    \hat{d} = d - \epsilon(d, \iangle),
\end{equation}
and corrected point $\hat{\mathbf{x}} = \mathbf{v} + \hat{d} \mathbf{r}$.
We evaluate the following depth correction models:
the polynomial model introduced in~\cite{Kuemmerle-2020-ICRA},
\begin{equation}~\label{eq:polynomial_model}
    \epsilon_\mathrm{p}(\iangle; w_1, w_2) = w_1 \iangle^2 + w_2 \iangle^4,
\end{equation}
and the polynomial model scaled by depth,
\begin{equation}~\label{eq:scaled_polynomial_model}
    \epsilon_\mathrm{sp}(d, \iangle; w_1, w_2) = d(w_1 \iangle^2 + w_2 \iangle^4),
\end{equation}
where $w_1$, $w_2 \in \reals$ are the learnable parameters of the models.
In the following, we will denote $\mparams$ the learnable parameters of the model
and $\hat{\s{X}}(\mparams)$ the set of points $\s{X}$ corrected using a model with parameters $\mparams$.

\subsection{Map Consistency Loss Functions}\label{sec:consistency_loss}
Let us denote
$\s{N}_{\s{X}}(\mathbf{x},r)$ the set of points $\mathbf{x}' \in \s{X}$ within distance $r$ from a given point $\mathbf{x}$.
For a set of points $\s{X}$, let us denote
$\bar{\mathbf{x}}$ the sample mean, $\bar{\mathbf{x}} = 1 / |\s{X}| \sum_{\mathbf{x} \in \s{X}} \mathbf{x}$, and
$\m{Q}$ the sample covariance matrix,
\begin{equation}\label{eq:cov}
\m{Q} = \frac{1}{|\s{X}| - 1} \sum_{\mathbf{x} \in \s{X}} (\mathbf{x} - \bar{\mathbf{x}}) (\mathbf{x} - \bar{\mathbf{x}})^\T.
\end{equation}
Note that $\m{Q}$ is a $3 \times 3$ real symmetric matrix.
We denote $0 \leq \lambda_1 \leq \lambda_2 \leq \lambda_3 \in \reals$ its eigenvalues and
$\mathbf{u}_1$, $\mathbf{u}_2$, $\mathbf{u}_3 \in \reals^{3 \times 1}$, $\|\mathbf{u}\| = 1$, $\mathbf{u}_i^\T\mathbf{u}_{j \neq i} = 0$, their corresponding eigenvectors, such that $\m{A} \mathbf{u}_i = \lambda_i \mathbf{u}_i$ for $i = 1$, \ldots, $3$.


In \emph{principal component analysis} (PCA), the eigenvectors of a sample covariance matrix are also referred to as principal components.
They form an orthogonal basis in which the individual dimensions of the data are linearly uncorrelated and which allows to form lower-dimensional projections of the data that preserve as much variance as possible in the given number of dimensions.
Eigenvalue $\lambda_i$ corresponds to the variance of the data projected onto the corresponding eigenvector $\mathbf{u}_i$.

We propose two distinct map consistency loss functions,
generically denoted $\ell$,
which are defined as functions of the sample covariance matrix $\m{Q}$ from (\ref{eq:cov}):
\begin{equation}\label{eq:min_eig_loss}
    \lambda_1\text{, its minimum eigenvalue, and}
\end{equation}
\begin{equation}\label{eq:trace_loss}
    \trace{\m{Q}}\text{, the sum of elements on its main diagonal.}
\end{equation}

The former corresponds to the variance along the direction in which the observed variance is minimum, i.e., along eigenvector $\mathbf{u}_1$.
Eigenvector $\mathbf{u}_1$ oriented toward the sensor origin $\mathbf{v}$ is also used as an estimate of surface normal vector, $\hat{\mathbf{n}} = -\sign{(\mathbf{u}_1^\T\mathbf{r})} \mathbf{u}_1$.
Since trace of a matrix corresponds to the sum of its eigenvalues, the latter loss function corresponds to the sum of variances along the three orthogonal directions given by eigenvectors $\mathbf{u}_i$, $\trace{\m{Q}} = \sum_{i = 1}^3 \lambda_i$.


One can see similarities with optimization criteria commonly used in iterative closest point (ICP) algorithms.
Point-to-plane ICP~\cite{Chen-1992-IVC} is similar to (\ref{eq:min_eig_loss}) in the sense that it only penalizes mis-alignment in the direction perpendicular to the surface
while point-to-point ICP~\cite{Besl-1992-TPAMI} penalizes mis-alignment in any direction, similarly to (\ref{eq:trace_loss}).
Unlike these and other ICP criteria~\cite{Segal-2009-RSS},
both proposed losses are formulated in terms of the resulting map and are thus independent of the order or the number of contributing input point clouds.
Moreover, the sampling points for which we gather the neighboring points, construct the covariance matrices, and compute the loss can also be chosen independently of input point clouds.

\subsection{Self-supervised Learning of Depth Correction}
In the following, a point $\mathbf{x} \in \reals^{3 \times 1}$ transformed by rigid transformation $\m{T}$ will be written simply as $\m{T}\mathbf{x}$ while a set of points $\s{X} \subset \reals^{3 \times 1}$ which are transformed by the same transformation will be written as $\m{T}\s{X}$.
Rigid transformation parameterized by vector $\mathbf{p} \in \reals^{6 \times 1}$ will be denoted by $\m{T}(\mathbf{p})$, here the pose vector $\mathbf{p} = [x, y, z, \boldsymbol{\theta}]$ contains position and orientation given in axis-angle representation $\boldsymbol{\theta} = \theta \mathbf{e}$.

\begin{figure*}
    \newcommand\imgwidth{0.5}
    \centering
    
    \subfigure[Points with enough neighbors, $\s{M}_{nbr}$~(\ref{eq:M_nbr}), $N_{min} = 10$]{\includegraphics[width=\imgwidth\columnwidth]{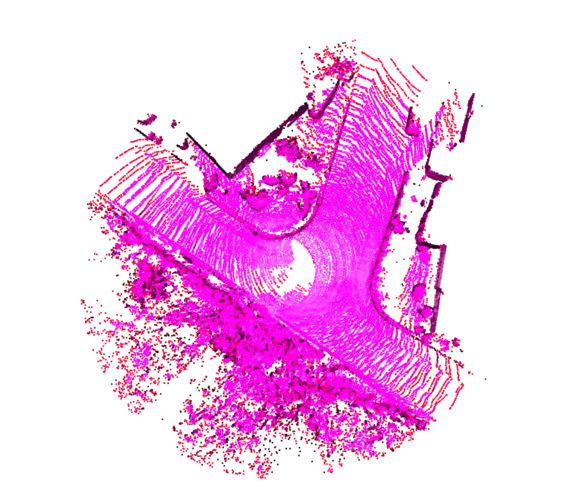}}
    \subfigure[Points belonging to flat surfaces, $\s{M}_{flat}$~(\ref{eq:M_plane}), $C_{0} = 0.25$]{\includegraphics[width=\imgwidth\columnwidth]{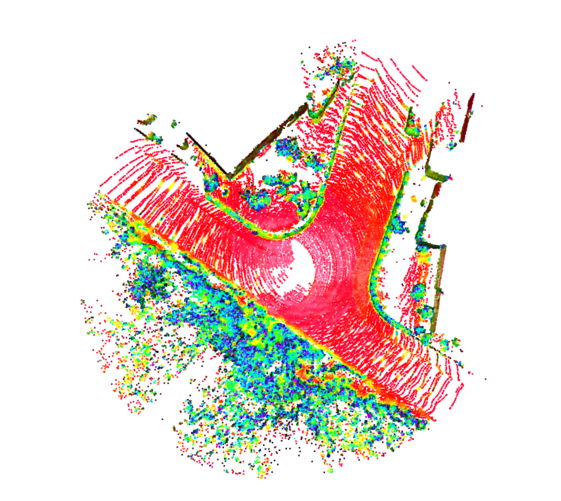}}
    \subfigure[View-points dispersion, $\s{M}_{disp}$~(\ref{eq:M_disp}), $\sigma_{min} = 0.36$]{\includegraphics[width=\imgwidth\columnwidth]{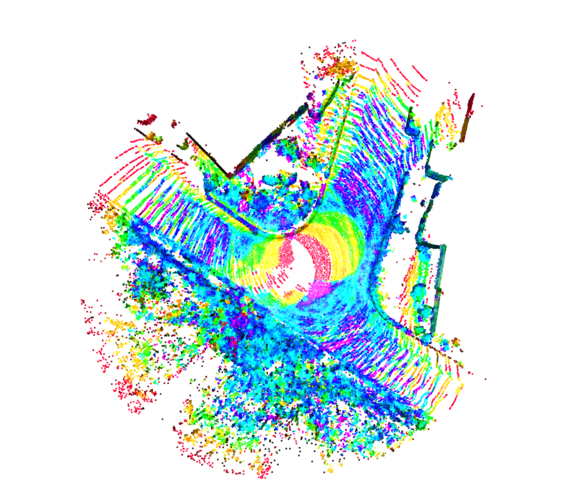}}
    \subfigure[Filtering result, $\s{M}_{\phi}~=~\s{M}_{nbr}~\cap~\s{M}_{flat}~\cap~\s{M}_{disp}$]{\includegraphics[width=\imgwidth\columnwidth]{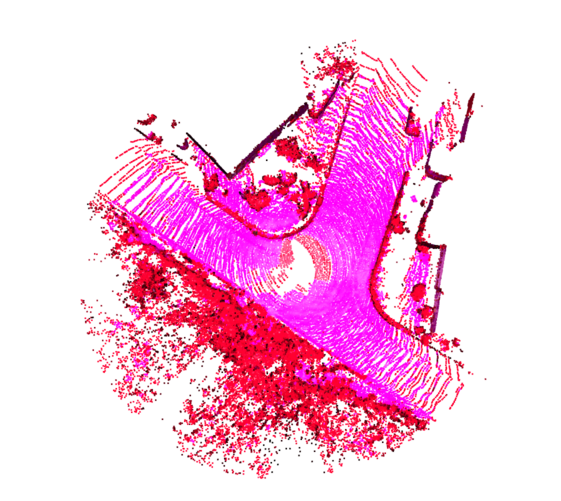}}
    
    \caption{Filters applied to select from a set of scans the points used in the optimization. Example from the KITTI-360 data set~\cite{Liao2022PAMI}.
    (a) Outliers with not enough neighboring points are shown in red. (b) Red points have a smaller eigenvalues ratio, $\frac{\lambda_1}{\lambda_2}$~(\ref{eq:M_plane}) (c) The rainbow color gradient from red to violet describes the points in the map observed from different sensor poses (view-points). Blue and violet points are measured from higher number of view-points. (d) Resulting filtering mask, purple points are selected in the optimization, Algorithm~\ref{alg:optimization}.}
    
    \label{fig:kitti360_opt_pts_filter}
\end{figure*}

The introduced map consistency losses can be used to learn parameters of the introduced depth correction models in a self-supervised way as follows.
First, we build global map $\s{M}$ from point clouds $\hat{\s{X}}_k$ corrected by a depth correction model and transformed into a global reference frame using transformations $\hat{\m{T}}_k$:
\begin{equation}\label{eq:map_contruct}
\s{M} = \bigcup_{k} \hat{\m{T}}_k\hat{\s{X}}_k(\mparams).
\end{equation}
Each transformation $\hat{\m{T}}_k = \m{T}_k \delta\m{T}(\mathbf{p}_k)$ is composed of
initial transform $\m{T}_k$, provided as ground truth or from SLAM,
and correction $\delta\m{T}(\mathbf{p}_k)$ parameterized by vector $\mathbf{p}_k \in \reals^{6 \times 1}$.
Second, we define a set of points that are used during the optimization using the projection $\phi: \s{M} \to \s{M}_{\phi}$ that selects a set of points that have \emph{enough neighbors}, belong to \emph{flat surfaces} and are observed from \emph{diverse} (several different) \emph{view-points}:
\begin{equation}
    \s{M}_{\phi} = \s{M}_{nbr} \cap \s{M}_{flat} \cap \s{M}_{disp}
    \label{eq:M_res}
\end{equation}
To put it formally, the individual subsets are defined as follows. The lower bound for the number of neighboring points is motivated by robust normal estimation.
\begin{equation} 
    \s{M}_{nbr} = \{ \mathbf{x}_i \in \s{M}: \s{N}_{\s{M}}(\mathbf{x}_i) \ge N_{min} \}
    \label{eq:M_nbr}
\end{equation}
As most of the points belonging to planar surfaces have a large incidence angle, we estimate the flat regions from the eigenvalue ratios.
\begin{equation} 
    \s{M}_{flat} = \{ \mathbf{x}_i \in \s{M}: \frac{\lambda_1}{\lambda_2} \le C_0, C_1 \le \frac{\lambda_2}{\lambda_3} \le C_2 \}
    \label{eq:M_plane}
\end{equation}
Additionally, in order to evaluate map consistency information, we need to use points measured from multiple sensor poses.
\begin{equation} 
    \s{M}_{disp} = \{ \mathbf{x}_i \in \s{M}: \sigma({\mathbf{p}(\mathbf{x}_i)}) \ge \sigma_{min}, \sigma(p) = \trace{\mathbf{\Sigma}_p} \}
    \label{eq:M_disp}
\end{equation}
Here the $\mathbf{\Sigma}_{p}$ denotes view-points covariance matrix and the $\sigma(\mathbf{p}(\mathbf{x}_i)$ is a view-point dispersion.
An example of the individual subsets selected from a point cloud map is given in~\autoref{fig:kitti360_opt_pts_filter}.

Third, we compute mean consistency loss $\ell$, (\ref{eq:min_eig_loss}) or (\ref{eq:trace_loss}), over selected set of points $\mathbf{x}_i \in \s{M_{\phi}}$,
\begin{equation}\label{eq:loss}
    L(\mparams, \mathbf{p}_1, \ldots, \mathbf{p}_\mathrm{K})
    = \frac{1}{|\s{M_{\phi}}|} \sum_{\mathbf{x}_i \in \s{M_{\phi}}} \ell(\mathbf{x}_i),
\end{equation}
from established neighborhoods $\s{N}_\s{M}(\mathbf{x}_i,r)$ and respective sample covariance matrices $\m{Q}_i$.
Forth, we compute the gradient of the loss with respect to the model parameters and individual transformation parameter vectors,
$\nabla_{\mparams} L(\mparams, \mathbf{p}_1, \ldots, \mathbf{p}_\mathrm{K})$ and
$\nabla_{\mathbf{p}_k} L(\mparams, \mathbf{p}_1, \ldots, \mathbf{p}_\mathrm{K})$,
using backpropagation in the computational graph.
Fifth, we update $\mparams$ and $\mathbf{p}_k$ using a gradient descent algorithm.
We repeat the above procedure for a fixed number of iterations and select the final model which minimizes the loss on a validation subset.
The neighboring sets $\s{N}_\s{M}(\mathbf{x}_i,r)$ computed for each point in global map, $\mathbf{x}_i \in \s{M}$, are fixed during the optimization loop.
\autoref{fig:depth_corr_scheme} describes map construction procedure, consistency loss computation, poses and depth correction model parameters updates (gradients back-propagation) from the Algorithm \ref{alg:optimization} which summarizes the described method.

\begin{algorithm}
\label{alg:optimization}
\caption{Depth Correction Optimization}

\KwInput{
    Point sets $\{\s{X}_k\}_{k = 1}^{\mathrm{K}}$,
    transformations $\{\m{T}_k\}_{k = 1}^{\mathrm{K}}$.
}
\KwOutput{
    Model parameters $\mparams$.
}

Initialize model parameters $\mparams \gets \m{0}$.

Initialize transformation corrections
$\mathbf{p}_k \gets \m{0}$ for $k = 1, \ldots, \mathrm{K}$.

\For{$i = 1, \dots, \mathrm{N}$}
{

Construct map according to (\ref{eq:map_contruct}),

$\s{M} \gets \bigcup_k \m{T}_k \delta\m{T}_k(\mathbf{p}_k)\hat{\s{X}}_k(\mparams)$.

Select valid map elements according to (\ref{eq:M_res}),
$\s{M}_{\phi} = \phi(\s{M})$.

Compute loss according to (\ref{eq:loss}),
$L(\mparams, \mathbf{p}_1, \ldots, \mathbf{p}_\mathrm{K}) \gets \frac{1}{|\s{M_{\phi}}|} \sum_{\m{x}_i \in \s{M_{\phi}}} \ell(\m{x}_i)$.

Compute loss gradients
$\nabla_{\mparams} L(\mparams, \mathbf{p}_1, \ldots, \mathbf{p}_\mathrm{K})$ and
$\nabla_{\mathbf{p}_k} L(\mparams, \mathbf{p}_1, \ldots, \mathbf{p}_\mathrm{K})$.

Update $\mparams$ and $\mathbf{p}_k$ using the computed gradients with a gradient descent algorithm.
}

Return $\mparams$ minimizing validation loss.
See text for the details.

\end{algorithm}












\section{Experiments and Results}~\label{sec:experiments}

We run a diverse set of tests with different criteria functions and depth correction models according to the scheme introduced in the Section~\ref{sec:methodology}.
A cross-validation method is used to split a data pull into training, validation and test sets.
The parameters (models' coefficients and poses corrections) are learned on training part.
The training performance is evaluated on validation data set.
The best models according to map consistency criteria are selected and their performance on testing part is recorded.
In the cross-validation scheme training, validation and testing data pulls are being swapped until all possible combinations are selected for the learning experiment.
This algorithm is applied to the experiments described in first row of the Table~\ref{tab:exps_overview}.
The questions of to which extent depth correction models remove measurement bias and how it affects localization quality in SLAM scenarios are addressed next (second row of the Table~\ref{tab:exps_overview}).
In this part, we compare SLAM pipelines with and without usage of the trained depth correction models. In the table $\mathbf{p}_i$ denotes ground-truth localization data, while $\hat{\mathbf{p}_i}$ are the estimated poses, Section~\ref{subsection:localization_acuracy}.
Data containing sequences of lidar scans and their corresponding locations is utilized.
We collect the data set in an indoor corridor environment, Section~\ref{subsection:dataset}. This choice is motivated by the presence of large planar surfaces.
It means that lidar beams fall at them at a wide spectrum of incidence angles which allows to observe the measurements bias.

\begin{table}
    \centering
    \caption{Experiments overview}
    \label{tab:exps_overview}
    \setlength{\extrarowheight}{3pt}
    \begin{tabular}[t]{p{19mm}|p{30mm}|p{23mm}}
        \hline
        name & purpose & estimated value \\
        \hline
        model learning and poses correction & learning of depth-correction model with
sensor poses refinement & $\mparams$, $\mathbf{\delta p}_1, \ldots, \mathbf{\delta p}_\mathrm{K}$ \\
        \hline
        localization accuracy & affect of depth correction models on localization accuracy & $\sum_{i=1}^{\mathrm{K}} \lVert \hat{\mathbf{p}}_i - \mathbf{p}_i \rVert_2 $ \\
        \hline

    \end{tabular}
\end{table}

\subsection{Lidar Bias Estimation}~\label{subsection:bias_estimation}

\begin{figure}
    \centering
    \includegraphics[width=\columnwidth]{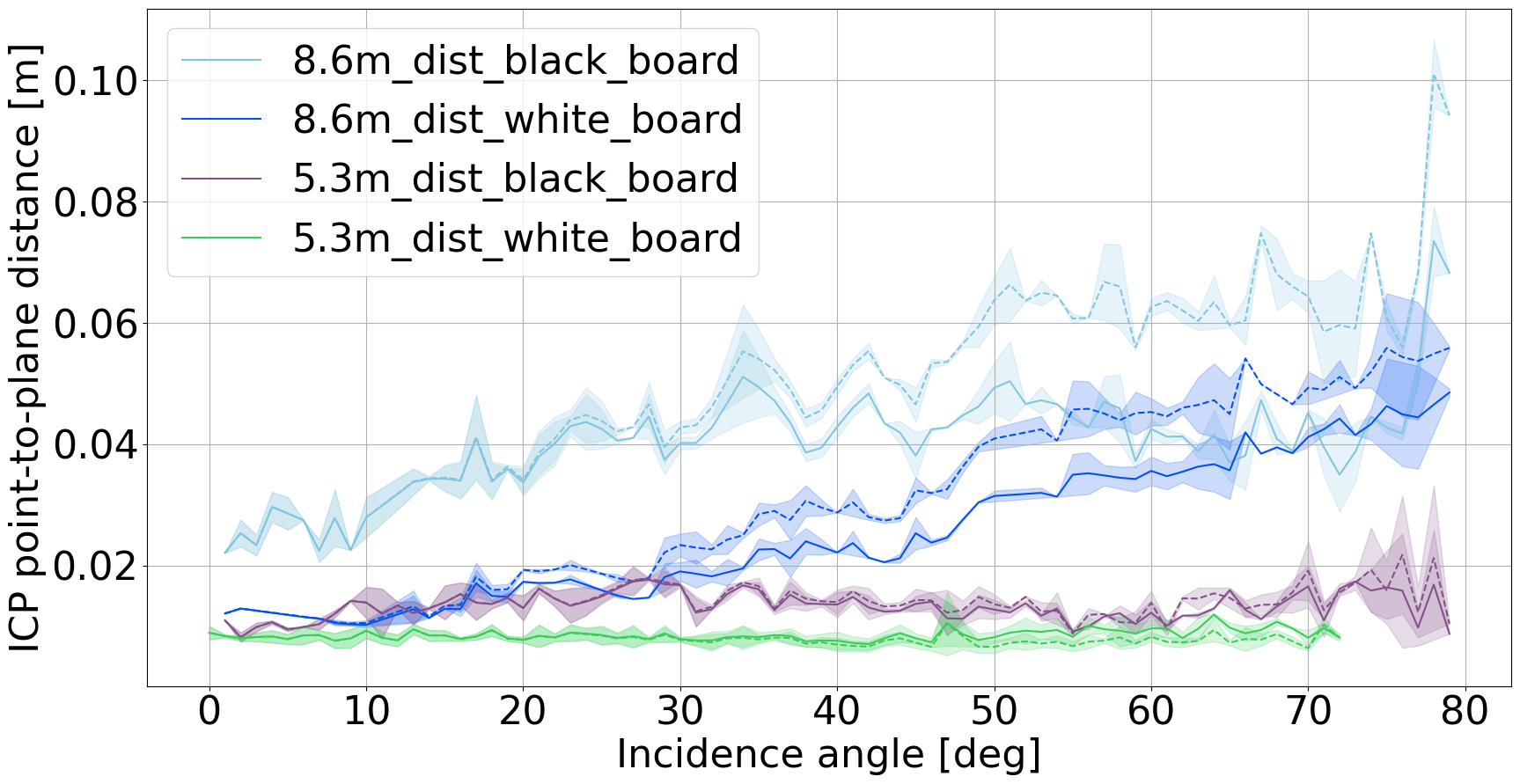}
    \caption{Dependence of the point-to-plane distance between measured cloud (by Ouster OS0-128 lidar) and ground-truth plane (given by the calibration board surface) on incidence angle. Results are provided for experiments with the board placed at $5.3 \si{\meter}$ and $8.6 \si{\meter}$ distance from the lidar. Dashed lines correspond to experiments without bias removal, while solid ones represent the error with depth correction applied to the same point clouds. For one experiment its own color is assigned.}
\label{fig:point-to-plane_inc_angle}
\end{figure}

\begin{figure*}
    \centering
    \subfigure[Rotated calibration board with known dimensions]{\includegraphics[width=1.3\columnwidth]{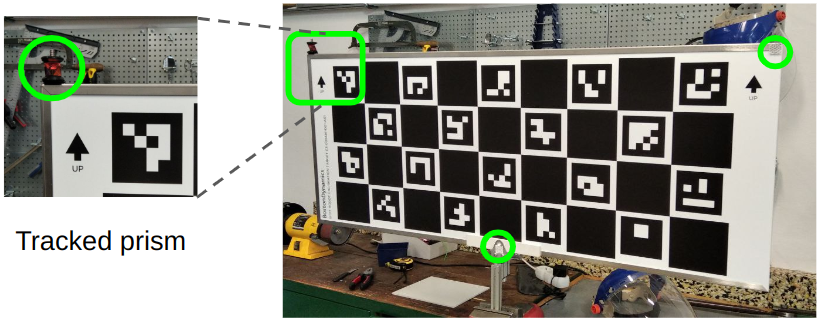}}
    \subfigure[Lidar point cloud and ground-truth board mesh.]{
    \includegraphics[width=0.6\columnwidth]{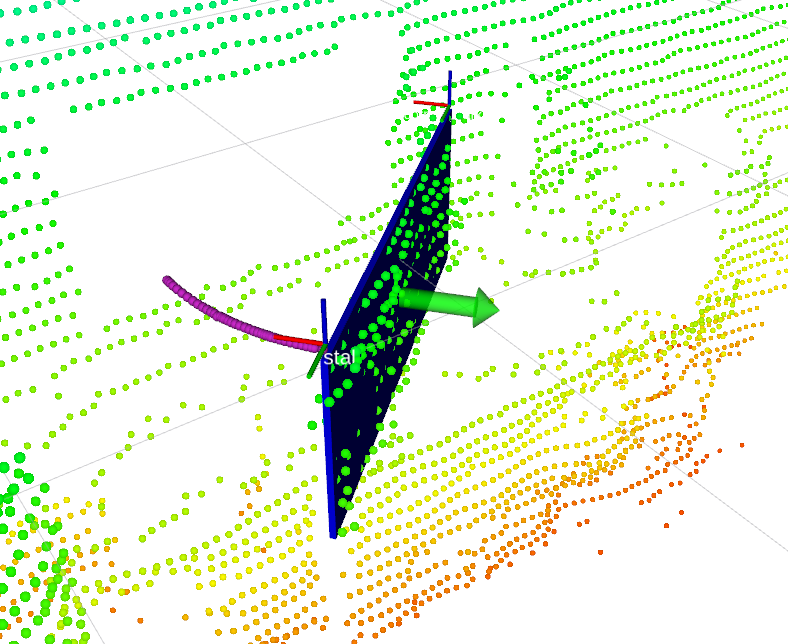}}
    \caption{Bias estimation experimental setup. \emph{Left}: the rigid board which could be rotated around vertical axis. The optical prism is attached to its corner. The prism is tracked by the theodolite. Points marked with green circles are measured with the theodolite once before running the experiment in order to precisely estimate board's dimensions. \emph{Right}: tracked corner trajectory is visualized in purple. Colored points represent lidar point cloud. The ground-truth board location is visualized as a blue mesh with green arrow denoting its normal direction.}
    \label{fig:bias_estimation_experiment}
\end{figure*}

The experimental setup to find out the dependence of lidar depth measurements on incidence angle is described in this section. We use the Ouster OS0-128 lidar~\footnote{hardware Rev C, \url{https://ouster.com/}}.
A rigid $0.50 \si{\meter} \times 1.15 \si{\meter}$ board with a metal frame is a target to measure with the lidar. The board could be rotated along the vertical axis~\autoref{fig:bias_estimation_experiment}.
The board position in space is measured both by the lidar and the theodolite Leica TS15.
As the theodolite provides measurements of one point in space, we place the optical prism to be tracked by the Leica~TS15 at the top left corner of the board. It provides the ground-truth location of the prism with $1 \si{\milli\meter}$ accuracy. The position of the lidar relative to the board is also measured by the theodolite.
In order to estimate the lidar orientation relative to the board's base (point of rotation), we run the ICP alignment of the point cloud sampled from the board mesh to the lidar point cloud at low angles of rotation of the board. The experimental board has two different surfaces: without marking (white one), and the calibration side with AR-markers drawn on it (black one). We run experiments with the board being placed at $5.3 \si{\meter}$ and $8.6 \si{\meter}$ distances from the lidar and measure both of its sides.
The goal is estimation of the difference between the measured board pose with respect to its ground truth for different orientations.
Knowing the board's dimensions and the its base is fixed, the full pose of the board is given. 
As the metric, we calculate the point-to-plane distance between the board's measured point cloud and the plane representing the board's surface scanned by the lidar.
The metric's values for different board angles are given in the~\autoref{fig:point-to-plane_inc_angle}. The error achieves $10 \si{\centi\meter}$ for incidence angles around $\ang{80}$ which confirms the results described in \cite{Laconte-2019-ICRA}. As dark surfaces have weaker reflective qualities, it was observed that the error is larger when the black side of the board was measured. The metric values for higher angles are not provided as there are almost no points measured by the lidar of the board oriented almost perpendicular. Note, that we did not observe dependence of the point-to-plane distance on incidence angles when the white-side target was placed reasonably close to the lidar (for $5.3 \si{\meter}$ distance).
Additionally, we provide the results corresponding to the point-to-plane value estimated for the lidar scans with the depth correction model (~\ref{eq:scaled_polynomial_model}) applied. They are depicted with the solid lines in the \autoref{fig:point-to-plane_inc_angle}.
Due to dependence on depth the same model provides larger corrections for more distant measurements ($8.6 \si{\meter}$).

\subsection{Data Set}~\label{subsection:dataset}

\begin{figure*}
    \centering
    \includegraphics[width=\textwidth]{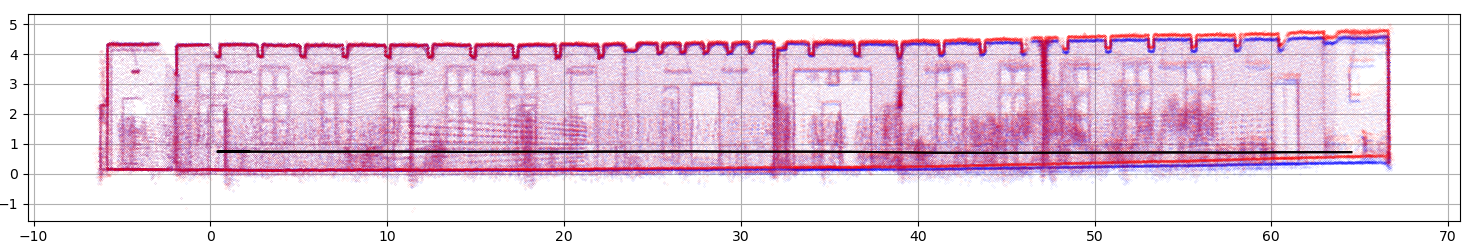}
    \caption{Side view of the corridor environment mapped in the data set~\ref{subsection:dataset}. Red point cloud denotes the map collected using baseline SLAM pipeline~\cite{Pomerleau-2013-AR}. Blue map was constructed using depth correction model to remove bias from lidar measurements before using the scans in SLAM alignment procedure. The black line corresponds to robot trajectory tracked with the theodolite. The distance scales along vertical and horizontal axes are chosen differently for illustration purposes.}
\label{fig:fee_corridor_map_bias}
\end{figure*}

\begin{figure*}
    \centering
    \subfigure[FEE Corridor data sequence]{\includegraphics[width=1.65\columnwidth]{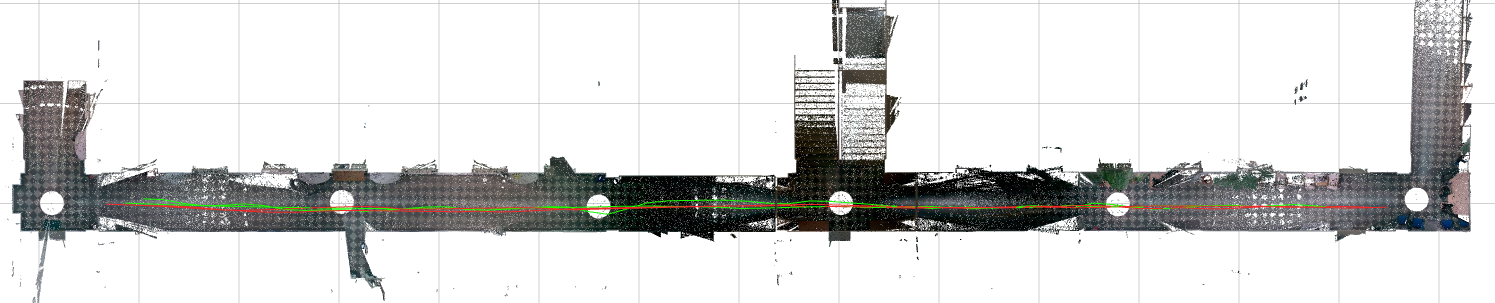}}
    \subfigure[Husky with lidar]{
    \includegraphics[width=0.35\columnwidth]{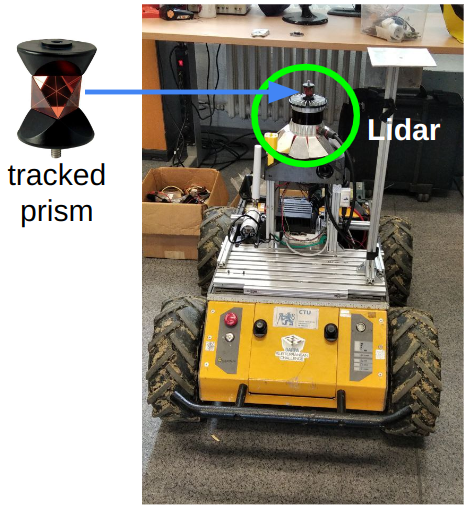}}

    \caption{Overview of the 3D model of the corridor (around $70 \si{\meter}$ long) environment generated with a Leica BLK360 scanner and the ground-truth trajectory obtained by the proposed approach. One cell of the grid is equal to 5 meters. The two trajectories traveled by the robot (corresponding to collected data sequences) are shown in red and green respectively. Husky robot with mounted Ouster OS0-128 lidar and tracked prism is shown on the right.}
    
    \label{fig:data_depth_correction}
\end{figure*}

In practice, however, it is not feasible to obtain exact positions and orientations of all the objects sensed by a lidar.
In order to train the depth correction models using the self-supervised scheme described in~\autoref{fig:depth_corr_scheme}, we collect the following data set (called "FEE Corridor").
It contains point clouds captured in an office indoor environment with precise localization and ground-truth mapping information~\autoref{fig:data_depth_correction}.
Two "stop-and-go" data sequences are provided of a robot with mounted Ouster OS0-128 lidar moving in the same environment.
This data-capturing strategy allows recording lidar scans that do not suffer from an error caused by sensor movement.
Individual scans from static robot positions are recorded.
Additionally, point clouds recorded with the Leica BLK360 scanner are provided as mapping ground-truth data.
The scanner is the tripod-mounted survey-grade lidar
utilized to capture a detailed millimeter-accurate 3D-maps.
Ground-truth 6~Degrees~of~Freedom (6 DoF) lidar poses are obtained using the SLAM introduced in \cite{Pomerleau-2013-AR} with alignment to the prerecorded ground-truth map.
In addition, the lidar poses are recorded with  the theodolite Leica TS15 tracker.
The theodolite tracks a prism placed on the robot.
In the corridor environment a direct line of sight of the prism is ensured for the whole time of recording.
The tracker provides translation data ($x_t$, $y_t$, $z_t$ coordinates of the prism relative to the tracker).
The tracked poses are also included in the data set for reference.
As we track only position of the prism, the orientation is provided by alignment of lidar scans to the pre-recorded with the Leica BLK360 scanner ground truth map.

As the importance of depth correction for autonomous cars was suggested already in the original work of Laconte \etal~\cite{Laconte-2019-ICRA},
we also train and assess depth-correction models on the data collected in the driving scenarios, namely on a part of the KITTI-360 benchmark~\cite{Liao2022PAMI}.
Here, the Velodyne HDL-64E sensor was used as point cloud measurements provider (at 10~\si{fps}), while ground-truth localization data
was estimated with GPS/IMU data fused with visual features.
We selected 8 sub-sequences each containing 50 consecutive scans (from 102 to 151) and corresponding lidar poses.
From these sub-sequences for depth correction training and validation we take every 5th scan and its pose due to computational reasons and to ensure the sufficient overlap between the neighboring sensor locations.
Sub-sequences of 500 consecutive (count from the beginning) scans were used in the evaluation of SLAM accuracy; see the Table~\ref{tab:loc_error_kitti360} describing the sub-sequences from KITTI-360 used for depth correction training and evaluation.


\subsection{Model Learning and Poses Correction}\label{subsec:results_pose_provider_gt}

In this experiment, the influence of depth correction models on map consistency is investigated.
We divide FEE Corridor sequences into 8 sub-sequences.
Each sub-sequence contains from 7 to 11 scans and corresponding lidar poses.
Next train (4 sub-sequences), validation (2 sub-sequences) and test groups (2 sub-sequences) are selected from them.
We utilize 4 cross-validation splits during training in order to have each sub-sequence in each group (train, validation or test).
The model with the best performance on validation set is saved during training.
Since ground-truth localization may be difficult to obtain due to the need for specialized hardware and usually a constrained stop-and-go operation,
in addition to learning models' parameters initial poses are being updated according to the optimization pipeline represented on~\autoref{fig:depth_corr_scheme}.
Note, that lidar position and orientation correction could be preformed differently based on the type of localization error one would like to eliminate.
In most of the real world cases, the sources of localization error are independent for individual poses at which laser scans were captured. Commonly, the transformations update would be computed as follow,
$\hat{\m{T}}_k = \m{T}_k \delta\m{T}_k$,
where $k$ denotes time index, Algorithm~\ref{alg:optimization}.
This poses update strategy was utilized in our work.
However, if the error of localization source could be considered negligible, it is still possible to encounter a bias connected with calibration inaccuracies (i.e. the error of a lidar location relative to the sensors suit origin).
It is reasonable to assume that the pose correction term is constant and independent on time at which a laser scan was taken, $\forall k = 1 \dots N: \delta \m{T}_k~=~\delta \m{T}$.

The trained models are used further in order to visualize the effect of measurement bias suppression qualitatively.
The point cloud map of the corridor environment is constructed with the help of the SLAM method~\cite{Pomerleau-2013-AR} without depth correction.
It is visualized in red on the~\autoref{fig:fee_corridor_map_bias}.
The lidar's ground-truth trajectory (recorded with Leica TS15 tracker) on the side view is depicted in black. The map is being banded towards the ceiling as the distance from the corridor origin increases.
We then run the same experiment with depth correction applied to raw lidar scans before being aligned in the SLAM algorithm. The resultant map is drawn in blue on the same~\autoref{fig:fee_corridor_map_bias}.
With the lidar measurements correction, the map drifts less at the end of the corridor.

\subsection{Localization Accuracy}~\label{subsection:localization_acuracy}




    
 

\begin{table}
    \caption{Localization accuracy on FEE Corridor data set.}\label{tab:loc_error}
    \centering
    \begin{tabular}{c | c | c}
    \hline
    method & $\Delta \boldsymbol{\theta}$ [\si{\deg}] & $\Delta \mathbf{t}$ [\si{\meter}] \\
    \hline

    baseline, SLAM \cite{Pomerleau-2013-AR} & 1.49 &	0.60 \\
    \hline

    SLAM \cite{Pomerleau-2013-AR} + DDD \cite{sterzentsenko2019self} & $7.26$ & $2.71$ \\
    \hline

    ours, SLAM \cite{Pomerleau-2013-AR} + $\epsilon_\mathrm{p}$ (\ref{eq:polynomial_model}) & $\boldsymbol{1.34}$ & $0.56$ \\
    
    ours, SLAM \cite{Pomerleau-2013-AR} + $\epsilon_\mathrm{sp}$ (\ref{eq:scaled_polynomial_model}) & $1.36$ & $0.56$ \\
 
    \hline
    \end{tabular}
\end{table}


\begin{table}
    \caption{Localization accuracy on KITTI-360 data set.}\label{tab:loc_error_kitti360}
    \centering
    \begin{tabular}{c|c|c|cc|cc}
        \hline
        \multicolumn{3}{c|}{method}& \multicolumn{2}{c|}{SLAM \cite{Pomerleau-2013-AR}} & \multicolumn{2}{c}{SLAM \cite{Pomerleau-2013-AR} + $\epsilon_\mathrm{p}$ (\ref{eq:polynomial_model})} \\
        \hline
        seq. & \makecell{start \\ scan} & \makecell{end \\ scan} &
        \makecell{$\Delta \boldsymbol{\theta}$ [\si{\deg}]} &
        \makecell{$\Delta \mathbf{t}$ [\si{\meter}]} &
        \makecell{$\Delta \boldsymbol{\theta}$ [\si{\deg}]} &
        \makecell{$\Delta \mathbf{t}$ [\si{\meter}]} \\
        \hline
         00 & 1 & 500 & 1.95 & 2.31  & \textbf{1.58} & \textbf{1.51} \\
         03 & 1 & 500 & 4.23 & 12.62 & \textbf{3.89} & \textbf{11.84} \\
         04 & 1 & 500 & 2.05 & \textbf{3.33} & \textbf{1.66} & 3.39 \\
         05 & 1 & 500 & 1.85 & 2.83  &  \textbf{1.37} & \textbf{2.19} \\
         06 & 1 & 500 & 2.21 & 3.34  &  \textbf{1.88} & \textbf{2.96} \\
         07 & 1 & 500 & 7.61 & 35.24 &  \textbf{7.46} & \textbf{34.18} \\
         09 & 1 & 500 & 4.04 & 12.16 &  \textbf{3.83} & \textbf{11.76} \\
         10 & 1 & 500 & 3.75 & 11.36 &  \textbf{2.84} & \textbf{8.91} \\
        \hline
    \end{tabular}
\end{table}

\begin{figure}
    \newcommand\imgwidth{0.48}
    \centering
    
    \subfigure{\includegraphics[width=\imgwidth\columnwidth]{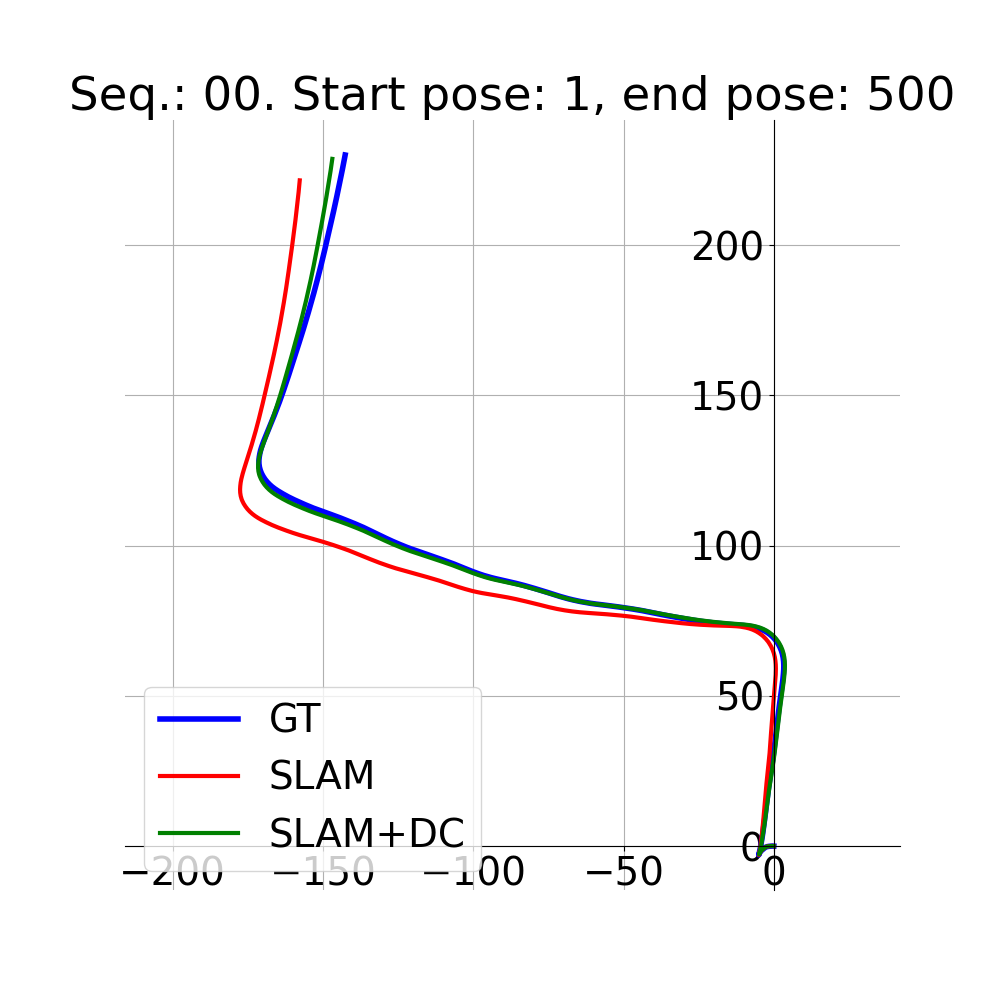}}
    \subfigure{\includegraphics[width=\imgwidth\columnwidth]{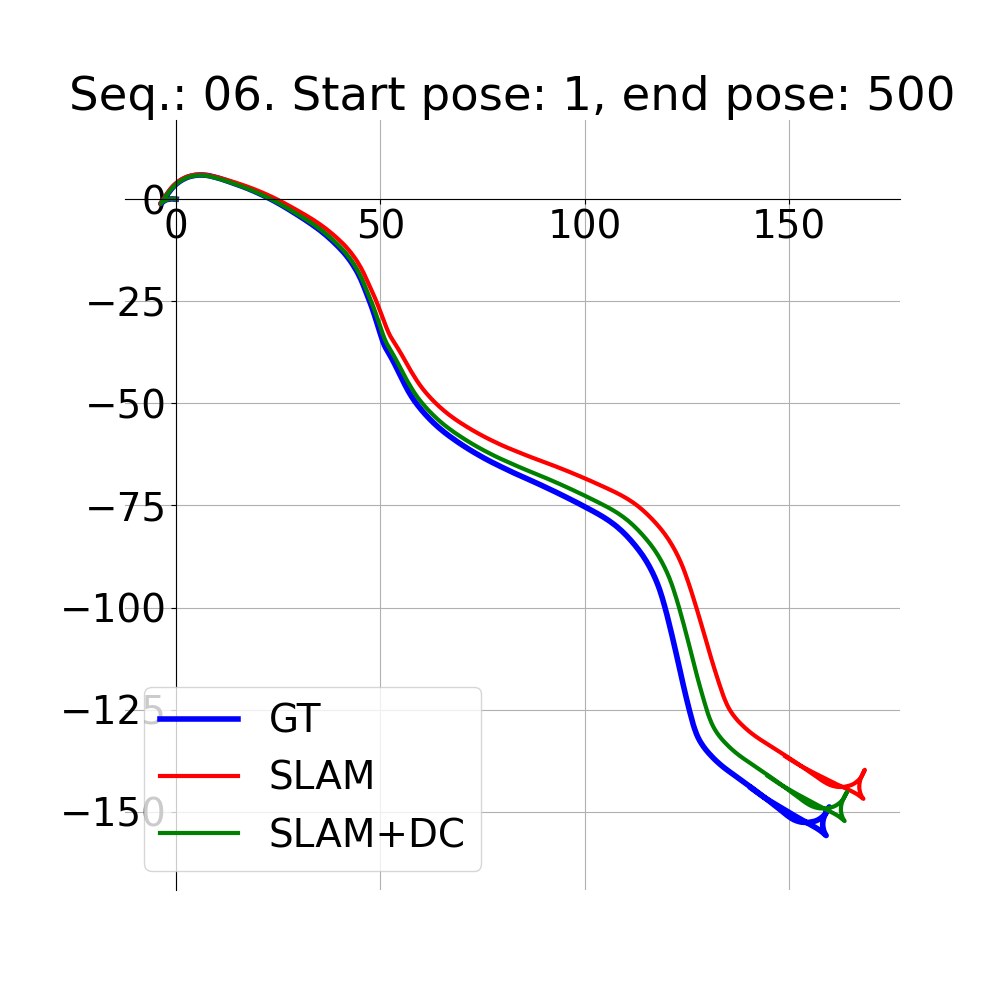}}
    
    \caption{Effect of depth correction on localization accuracy. The two examples from KITTI-360 data set are shown: the first 500 poses from the sequences $04$ and $06$ are given. \textit{Blue} curves represent ground-truth trajectories, \textit{red} ones are produced by the SLAM~\cite{Pomerleau-2013-AR}, the \textit{green} ones correspond to \textbf{our} pipeline: SLAM~\cite{Pomerleau-2013-AR} + $\epsilon_{p}(\ref{eq:polynomial_model})$. See Table~\ref{tab:loc_error_kitti360} for more details.}
    
    \label{fig:kitti360_slam_dc_results}
\end{figure}

Up to this point we made sure, that the depth correction models help to create more accurate point cloud maps by elimination of the bias related to measuring objects with high incidence angle.
The next reasonable question is, "Once we have more accurate lidar measurements, would it be possible to improve the localization accuracy of the sensor?".
In order to address it, we benchmark the SLAM method, \cite{Pomerleau-2013-AR} on the sequences from the data set with the ground-truth localization information.
In the first part of the experiment, the SLAM pipeline was running using the sensor measurements available in data sequences directly.
Then we execute the same procedure with the difference: depth correction is applied to lidar scans before the SLAM point cloud alignment operation.
Additionally, we run the similar experiment with the Deep Depth Denoising (DDD) method introduced in the~\cite{sterzentsenko2019self} as the depth correction part.
As the DDD model was trained with the RGB-D data, in the experiments point clouds are transformed to range images. The vertical resolution of an image is defined by the lidar's number of beams.
The depth denoising operation is applied only for points in distance range (up to $6.5 \si{\m}$) that was used by the authors of~\cite{sterzentsenko2019self} during the model training.
For each version of the experiment, the localization error is computed and averaged over time.
We split the ground-truth trajectories from the FEE Corridor into poses uniformly spread $1 \si{\meter}$ apart and record their corresponding time moments.
We then evaluate the poses from SLAM at the same time moments and compare them to ground-truth poses.
The results for depth correction models trained with minimum eigenvalue loss (\ref{eq:min_eig_loss}) are given in the Table~\ref{tab:loc_error} ($\Delta \boldsymbol{\theta}$ is orientation error, $\Delta \mathbf{t}$ is translation error).
It is demonstrated that removing the bias from lidar scans improves localization accuracy by around $7\% (4 \si{\centi\meter})$ in translation and $9\% (\ang{0.15})$ in orientation. For the DDD model experiments, we observe a significant localization error mostly caused by a drift in a vertical direction (more than $2.5 \si{\m}$ along $z$-axis only).
We perform similar experiments on the KITTI-360 data set, refer to the Table~\ref{tab:loc_error_kitti360}. It can be noticed, that the suppression of the depth bias from lidar measurements decreases both translation and orientation errors for all except one (number $04$) sequences used in the experiments. \autoref{fig:kitti360_slam_dc_results} contains examples of the SLAM pipeline runs on the first 500 scans from the sequences $00$ and $06$.
The correction of lidar measurements takes around $80 \si{\milli\second}$ (including normals and incidence angles estimation) on a modern CPU.
The implementation bottleneck lies in a point cloud normals estimation that could be overcome by GPU-based sparse tensor voting \cite{liu2012normal}.
We run the SLAM evaluation experiments offline, however, the implementation speed-up is considered as a topic for the future work that will allow to use the module in real-time scenarios.

\section{Conclusion}~\label{sec:conclusion}

In this paper, novel point cloud map consistency losses as well as depth measurement correction models were presented.
To tackle the lack of ground-truth depth data, the models were trained using lidar scans taken at multiple viewpoints of the same scenes in a self-supervised manner.
Once the models were  trained, it was shown that with their help the correction of lidar scans allows the construction of a higher-quality point cloud map.
It was also demonstrated that depth correction reduces the drift of lidar localization.
In addition to that, we release the data set with accurate ground-truth 3D-poses and mapping information for point cloud registration algorithms.







\bibliographystyle{plain} 
\bibliography{references.bib}

\end{document}